\documentclass[sn-mathphys-num]{sn-jnl}

\usepackage[caption=false,font=normalsize,labelfont=sf,textfont=sf]{subfig}
\usepackage{textcomp}
\usepackage{stfloats}
\usepackage{url}
\usepackage{booktabs}
\usepackage{dirtytalk}
\usepackage{multirow}%
\usepackage{amsthm}%
\usepackage{mathrsfs}%
\usepackage[title]{appendix}%
\usepackage{xcolor}%
\usepackage{manyfoot}%
\usepackage{listings}%
\usepackage{amsmath,amsfonts}
\usepackage{algorithmic}
\usepackage{array}
\usepackage{graphicx}

\theoremstyle{thmstyleone}%
\theoremstyle{thmstyletwo}%

\theoremstyle{thmstylethree}%

\begin{document}

\title[Article Title]{Tec-Habilidad: Skill Classification for Bridging Education and Employment}


\author*[1,2]{\fnm{Sabur} \sur{Butt}}\email{saburb@tec.mx}
\author[1]{\fnm{Hector G.} \sur{Ceballos}}\email{ceballos@tec.mx}
\author[2]{\fnm{Diana P.} \sur{Madera}}\email{A01025835@tec.mx}

\affil*[1]{\orgdiv{Institute for the Future of Education}, \orgname{Tecnológico de Monterrey}, \orgaddress{\city{Monterrey}, \postcode{464849}, \country{Mexico}}}

\affil[2]{\orgdiv{School of Engineering and Sciences}, \orgname{Tecnológico de Monterrey}, \orgaddress{\city{Estado de México}, \postcode{52926}, \country{Mexico}}}


\abstract{Job application and assessment processes have evolved significantly in recent years, largely due to advancements in technology and changes in the way companies operate. Skill extraction and classification remain an important component of the modern hiring process as it provides a more objective way to evaluate candidates and automatically align their skills with the job requirements. However, to effectively evaluate the skills, the skill extraction tools must recognize varied mentions of skills on resumes, including direct mentions, implications, synonyms, acronyms, phrases, and proficiency levels, and differentiate between hard and soft skills. While tools like LLMs (Large Model Models) help extract and categorize skills from job applications, there's a lack of comprehensive datasets for evaluating the effectiveness of these models in accurately identifying and classifying skills in Spanish-language job applications. This gap hinders our ability to assess the reliability and precision of the models, which is crucial for ensuring that the selected candidates truly possess the required skills for the job. In this paper, we develop a Spanish language dataset for skill extraction and classification, provide annotation methodology to distinguish between knowledge, skill, and abilities, and provide deep learning baselines to advance robust solutions for skill classification.}




\maketitle
\section{Introduction}

In an era marked by rapid technological advancement and dynamic labor market demands, the ability to accurately identify and categorize skills from textual data has become increasingly valuable. AI-based skill classification contributes to educational foresight, allowing institutions to anticipate future workforce demands and adjust curricula accordingly. With industries increasingly relying on data-driven decision-making, AI-powered insights can help educational institutions identify skill gaps, predict emerging job roles, and ensure that students develop competencies aligned with evolving economic landscapes. This proactive approach enhances not only employability but also the overall sustainability of the education system. 

Resumes, job descriptions, syllabi, performance reviews, online portfolios, industry reports, and many other resources where skills are mentioned are all textual in modality~\cite{khaouja2021survey}. Going through these texts manually is cumbersome, bias-inducing, and time-consuming, especially in large corporate industries where several skill-based assessments and analyses are needed every day. Hence it is crucial to develop automatic skill assessment technologies and algorithms that accurately identify skills and later classify them into different taxonomies for analysis. Skill assessment is a systematic process that involves several key steps to evaluate and understand the competencies of individuals or groups. This process includes multiple layers including skill identification~\cite{khaouja2021survey}, skill classification~\cite{putka2023evaluating, zhang2022skillspan}, skill mapping~\cite{mittal2019skill}, and more.

A significant gap exists in the field of natural language processing for skill identification and classification in Spanish. While there are numerous datasets available for English~\cite{zhang2022skillspan, khaouja2021survey}, the resources for Spanish, particularly in the context of skill identification and classification, are limited. This vacuum hampers the development and application of advanced NLP models for the Spanish-speaking job and educational market, where understanding and categorizing skills accurately is crucial for effective human resource management, career development, job matching, and school curriculum alignment~\cite{ali2023edguide}. The dataset for context-based skill identification and automatic skill classification in Spanish fills a critical gap in understanding the skills and classifying them based on an internationally accepted taxonomy. \textit{We developed the first Spanish-based skill identification and classification dataset.}

One effective method for classifying skills is through the Knowledge, Skill, and Abilities (KSA) taxonomy. This framework categorizes competencies into three distinct areas: knowledge, which encompasses theoretical understanding and information; skills, which involve the practical application of knowledge and abilities to perform tasks; and abilities, which refer to the innate or developed capacities to carry out activities. Using the KSA taxonomy provides a structured approach to skill identification, helping to clarify the specific attributes required for different roles and improving the precision of classification systems. \textit{Our developed classification dataset classifies each extracted skill, into a Knowledge, Skill, Abilities, or Other (KSAO) class. The \say{Other} class is important as it helps identify all the extracted requirements/rules that do not fall into the KSA category.}

Along with reliable datasets, we also require reliable classification methods for skill classification. Historically, Natural Language Processing (NLP) researchers have leaned on both unsupervised and supervised approaches, encompassing techniques~\cite{zhao2015skill, zhang2022skillspan, kivimaki2013graph} like skill embeddings, LSTM, BERT, among others. Unsupervised methods such as topic modeling necessitate interpretation by domain experts, adding a time constraint and hindering scalability across various sectors. Moreover, these approaches might struggle with contextual understanding, potentially leading to incomplete skill identification. Similarly, primitive supervised learning techniques lack recent skill identification and are not exposed to the model in the training phase due to a lack of contextual understanding. Hence, we leverage large language models and transformer methods for skill identification and classification to tackle these issues. \textit{Our findings indicate that transformer-based models can achieve high accuracy in skill identification tasks, surpassing the inter-annotator agreement levels.}

Our paper analyzes the following research questions: (\textbf{RQ1}) Can Large Language Models accurately predict hard and soft skills using job descriptions in a zero-shot setting? We explain the complete skill extraction and validation methodology in Section~\ref{Annotation Guidelines} and ~\ref{Section: Agreement}. (\textbf{RQ2}) How accurately can deep learning techniques predict KSAO (Knowledge, Skills, Abilities, and Other) labels? Answering this question, we explain the performance of our models in Section~\ref{Section: Results}. (\textbf{RQ3}) What are the challenges in correctly labeling and predicting KSAOs? Where does the model fail? We answered this question in detail in Section~\ref{Section: Error Analyses} of the paper.


\section{Literature Review}

\subsection{Interface between Education and Employement}
One of the major challenges in workforce development is the disconnect between what is taught in academic institutions and what is required in the labor market~\cite{bassi2012disconnected}. Many graduates find themselves underprepared for industry roles due to the lack of practical, job-relevant skills. According to previous research~\cite{goulart2022balancing}, skill identification through automated methods can facilitate more precise curriculum design, ensuring that students acquire competencies that align with job market trends. Moreover, skill classification frameworks~\cite{caratozzolo2024novel,gonzalez2024analyzing}, such as the Knowledge, Skills, and Abilities (KSA) taxonomy, provide a structured approach to mapping educational content to industry expectations. The increasing use of technology in skill extraction has also demonstrated its value in mitigating hiring biases and enhancing the efficiency of recruitment processes. By leveraging Natural Language Processing (NLP) techniques~\cite{gonzalez2024analyzing}, organizations can automatically extract, classify, and map skills from resumes, job descriptions, and academic records. This allows for a more objective evaluation of candidates and helps job seekers understand the specific competencies they need to develop.

As industries undergo digital transformation, workers need continuous learning opportunities~\cite{vaataja2024needs,li2024reskilling} to remain competitive. Reskilling and upskilling initiatives are essential to help individuals transition between roles or adapt to new technologies. However, the effectiveness of these initiatives depends on accurate skill identification and classification. Organizations and educational institutions must identify existing skill gaps, determine the relevant skills required for emerging jobs, and offer targeted training programs. Recent advancements in skill extraction and classification using deep learning models and Large Language Models (LLMs)~\cite{gonzalez2024analyzing,herandi2024skill} have shown promising results in improving the accuracy of skill identification. These models can distinguish between hard skills (technical competencies) and soft skills (interpersonal and cognitive abilities), ensuring a more comprehensive understanding of workforce capabilities. Furthermore, by utilizing a structured taxonomy such as KSAO (Knowledge, Skills, Abilities, and Other), organizations can systematically track employee progress and tailor upskilling programs to meet specific industry needs.

\subsection{Skill Extraction}
The skill extraction (SE) field has seen various approaches, primarily differentiated by their annotation levels (document, sentence, span), sources of labels (crowdsourcing, domain experts, predefined inventories), and the types of skills targeted (hard, soft, or both). Early works such as~\citet{kivimaki2013graph} and~\citet{zhao2015skill} employed document and sentence-level annotations using models like LogEnt and Word2Vec for hard skills extraction. More recent studies, such as~\citet{sayfullina2018learning} and~\citet{bhola2020retrieving}, introduced span-level annotations and utilized deep learning models, including CNNs, LSTMs, and BERT, often focusing on either soft skills or a predefined set of skills. 

Neural network models have demonstrated effectiveness in identifying transversal skills in job ads, even with limited data, thanks to their ability to detect complex patterns. However, the datasets often face challenges due to a significant imbalance between skill-containing sentences and those without, as most job ads do not explicitly reference transversal skills. To address this, authors such as~\citet{leon2024hierarchical} used a data augmentation technique involving sentence cloning to replicate sentences with skills, balancing the dataset and enhancing the models' robustness and accuracy. However, a related challenge with neural networks is the complexity of training as involves navigating a vast search space, including decisions about network architecture and hyperparameters, with the large number of possible configurations making exhaustive exploration impractical.

Conversely, the SKILLSPAN dataset, as introduced by~\citet{zhang2022skillspan}, addresses several gaps identified in previous studies. It offers span-level annotations for both hard and soft skills, annotated by domain experts, which is a significant departure from the more common crowd-sourced or inventory-based labels. The dataset includes 14.5K sentences with over 12.5K annotated spans, providing a comprehensive resource for SE tasks. The authors also release detailed annotation guidelines, enhancing the reproducibility and transparency of their work. The authors establish a BERT baseline and experiment with language models optimized for long spans, such as SpanBERT, JobBERT, and JobSpanBERT: Variants of BERT and SpanBERT, respectively, further pre-trained on a corpus of job postings to adapt to the specific domain. This model consistently performed the best across multiple tasks and datasets. Specifically, on the test set for predicting knowledge, JobBERT achieved an F1 score of 63.88\%, which was the highest among all models and configurations tested. We took a different approach for skill extraction, using the extractive capabilities~\cite{goel2023llms} of newly emerging Large Language Models (LLMs) that have been seen to perform exceedingly well~\cite{farooq2023chatgpt} on zero-shot and finetuned datasets. Since we are using Instruction-tuned LLMs, we also don't have to rely upon huge amounts of annotated datasets. 

Nowadays, some state-of-the-art approaches using LLMs follow methods similar to those outlined by~\citet{clavié2023large}, who describes a two-step approach to skill matching. First, linear regression classifiers are applied to frozen embeddings, and cosine distance between embeddings is measured to generate an initial list of potential skill matches. In the second step, LLMs are used as zero-shot rerankers to re-evaluate and refine this list. While LLMs are commonly recognized for their few-shot learning capabilities, research such as~\citet{kojima2023large} indicates that they can also perform surprisingly well in zero-shot scenarios when guided to think step-by-step. Consequently, despite not being trained on annotated skill-matching data, LLMs effectively rerank the matches, demonstrating their capacity to handle such tasks with minimal training data.

Furthermore,~\citet{nguyen2024rethinking} assesses the effectiveness of Large Language Models across six publicly available SE datasets, highlighting how the extracted skills contribute to a diverse and comprehensive taxonomy. The use of LLMs improves the understanding of the context in which skills are mentioned, manages variations and synonyms more effectively, and adapts to different industries and domains, leading to more flexible skill identification. Despite certain limitations, such as the tendency to split compound skills into separate entities, the authors propose that these skills can be refined and classified by domain experts to improve the taxonomy's accuracy and granularity.

\subsection{KSA taxonomies}
This refinement process is important in the context of a knowledge base, defined as a collection of records stored in a database, that typically encapsulates information about various aspects of the world. Prior research on skill identification from job advertisements, focusing on the skill bases employed, categorizes these into predefined and customized skill bases. Predefined skill bases, developed by experts, include ESCO (European Skills/Competences, Qualifications, and Occupations), O*NET (Occupational Information Network), ROME, KSA (Knowledge, Skill, and Abilities), and ISCO-8 (International Standard Classification of Occupations)~\cite{khaouja2021survey}. Classification of these skills based on the taxonomies gives us deeper insights into how the industry and academic sectors are shaping similarly or differently. 

However, as stated by~\citet{anelli2023} despite being available in multiple languages and covering a diverse range of countries with varying levels of development, the non-English dictionaries are generally smaller compared to their English counterparts in predefined bases like ESCO. This discrepancy can lead to gaps in effectively capturing the nuances of local labor markets. The Chilean study highlighted this issue, revealing significant differences when the ESCO dictionary was supplemented with a manually constructed dictionary. This custom dictionary was tailored to reflect the specific vocabulary used by employers in Chilean labor markets, demonstrating the importance of localizing and expanding such dictionaries to better align with regional linguistic and economic contexts.

Understanding how effectively education and training systems equip individuals with the necessary skills and address skill mismatches requires a well-defined framework and robust indicators. These elements are crucial for achieving positive labor market and social outcomes. As noted by~\citet{rodrigues2021aunified}, challenges such as confusion and inconsistency in the concepts used in policy documents and research related to skills and competencies can be addressed by providing clear definitions and establishing connections that integrate the analysis of skill supply and demand in both education systems and the labor market. To this end, we propose using a KSAO taxonomy for developing a Spanish-based skill identification and classification dataset.

Exploring the potential of training a machine to accurately~\citet{putka2023evaluating} estimate Knowledge, Skills, Abilities, and Other characteristics (KSAO) ratings for jobs using job descriptions and task statements as inputs, the authors utilized data from the U.S. Department of Labor’s O*NET system for 963 occupations and an independent dataset from a large organization for 229 occupations, the researchers found that their approach produced KSAO predictions with cross-validated correlations to subject matter expert (SME) ratings .74, .80, .75, and .84 for knowledge, skills, abilities, and others, respectively. The validity of these machine-based predictions was demonstrated through convergence with SME ratings, meaningful patterns in regression coefficients, and the conceptual relevance of predictor models. This solidifies our efforts to create datasets for supervised skill identification and classification task methodologies. 

\section{Dataset}
The data consists of extracted job postings from Indeed Mexico~\footnote{\url{https://mx.indeed.com/}} for December 2023 to January 2024. We extracted the job description, job ID, URL, job title, salary, employment type, and job location. The job postings were from the automotive industry in Mexico. For skill identification, we used the job description and extracted all the possible skills mentioned in the job posting. The data was sourced exclusively from seven prominent companies operating in Mexico, namely BMW, General Motors, Toyota, Kia, Nissan, Tesla, and Volkswagen. These companies were selected based on their significant presence in the Mexican market. It is noteworthy that all data collected pertains solely to job postings within Mexico. Table~\ref{tab:example_skillidentification}, and~\ref{tab: skill_classification_examples} show the dataset's structure and examples. The dataset is available publicly on request at DataHub link~\cite{FK2/O7E66L_2025} for research purposes.

\begin{table*}[!hbtp]
\caption{Example from the skill identification dataset with gold skill and soft skill labels\label{tab:example_skillidentification}}
\resizebox{\textwidth}{!}{%
\begin{tabular}{@{}lllllllll@{}}
\toprule
Title  & Salary & Description & Employement type & Job location & Skill & Soft skill \\ \midrule

\begin{tabular}[c]{@{}l@{}}Valuador \\automotriz\end{tabular}   
&   \begin{tabular}[c]{@{}l@{}} \$18,000 a\\  \$20,000 \\ por mes\end{tabular} &   \begin{tabular}[c]{@{}l@{}}BMW Grupo Carmen Motors lider\\ en el ramo automotriz premium\\ busca Valuador de autos seminuevos.\\ 
Perfil: Nivel de estudios: Licenciatura\\ en Administracion de Empresas, Comercio,\\ Negocios Internacionales o afines.\\ Conocimientos: excel, compra de unidades,\\ valuaciones. 
Competencias: enfoque \\en resultados, organizacion, analisis, \\negociacion y liderazgo.
\\Experiencia: en seminuevos en\\ distribuidores autorizados. No lotes.
\\Funciones principales: Realizacion\\ de avaluos de unidades.\\
Organizacion de reportes de\\ valuaciones y medicion constante\\ de los mismos. Cuantificar\\ de manera exacta costos reales\\ de reacondicionamiento.\\ Estimar, cuantificar y valorar las unidades...    \end{tabular}      
&    \begin{tabular}[c]{@{}l@{}}Tiempo\\ completo   \end{tabular}            
&     \begin{tabular}[c]{@{}l@{}}  Agencia \\BMW Carmen\\Motors avenida\\ Patria
Zapopan, Jal.  \end{tabular}     
& \begin{tabular}[c]{@{}l@{}} 1)Conocimientos en excel\\
2)Compra de unidades\\
3)Valuaciones\\
4)Experiencia en seminuevos\\en distribuidores autorizados\\
5)Realizacion de avaluos de unidades\\
6)Organizacion de reportes de\\valuaciones\\
7)Medicion constante de los reportes\\
8)Cuantificacion de costos de\\reacondicionamiento\\
9)Estimacion, cuantificacion y\\valoracion de unidades\\
10)Toma y compra de unidades \end{tabular}       
&   
\begin{tabular}[c]{@{}l@{}}
1)Enfoque en resultados\\
2)Organizacion\\
3)Analisis\\
4)Negociacion\\
5)Liderazgo   \end{tabular}     \\ \bottomrule
\end{tabular}}

\end{table*}

\begin{table}[!hbtp]
\caption{Example from the raw skill classification dataset with gold labels.\label{tab: skill_classification_examples}}
\begin{tabular}{@{}ll@{}}
\toprule
Skills & Label \\ \midrule
\begin{tabular}[c]{@{}l@{}}Promedio academico minimo de \\80 (Minimum academic average of 80)   \end{tabular}   &   Other    \\
\begin{tabular}[c]{@{}l@{}}Conocimientos en mecanica, electricidad, neumatica e hidraulica \\(Knowledge of mechanics, electricity, pneumatics and hydraulics) \end{tabular}     &    Knowledge   \\

\begin{tabular}[c]{@{}l@{}}Realizacion de compras \\ (Making purchases) \end{tabular}     &    Skill   \\

\begin{tabular}[c]{@{}l@{}}Auto confianza \\ (Self-confidence) \end{tabular}     &    Ability   \\

\bottomrule
\end{tabular}%

\end{table}

\subsection{Data collection}
We initiated the data collection process by developing a custom web scraper tailored to extract job postings and their corresponding descriptions from Indeed Mexico. This scraper was designed to navigate the Indeed platform and retrieve relevant data efficiently. The data collection period spanned from December 2023 to January 2024, ensuring that the gathered information encapsulates recent job postings within the specified timeframe. This temporal scope was chosen to reflect the most current trends and requirements in the job market. To ensure consistency and relevance within the context of the Mexican job market, all extracted data (original job descriptions) were translated from English to Spanish. Moreover, all subsequent analysis and processing were conducted exclusively in Spanish, maintaining the integrity and coherence of the collected information within the local linguistic context.

\subsection{Annotator Details}
We recruited 3 annotators, 2 male and 1 female who were computer science students at a private Mexican university. They were given a detailed annotation guide as well as a training workshop before the annotation process. 

\subsection{Annotation Guidelines} \label{Annotation Guidelines}

\subsubsection{Skill Identification}
In the skill identification dataset, annotators were tasked with aligning skills and soft skills generated by GPT-3.5-Turbo~\footnote{\url{https://platform.openai.com/docs/models}} with job descriptions to assess their relevance. Measures were taken to ensure there was no redundancy or overlap of keywords in the respective columns. To verify the accuracy of skill identification, we took the subset of the entire dataset including the job postings from General Motors, BMW, and Tesla. The annotator was told to keep only the skills that were written verbatim in the job descriptions and any abstract skills to be removed for the verification, making it a skill extraction task. 

\subsubsection{Skill Classification}
For the skill classification dataset, we followed~\cite{caratozzolo2023matrix, conklin2005taxonomy, hoque2016three} work in differentiating Knowledge, Skills, and Abilities. The annotators were required to label null or others as additional tags that were neither skills, abilities, or knowledge. The following key details were used to separate Skill, from Knowledge and Abilities:
\begin{itemize}
    \item Knowledge: Knowledge is defined as comprehension, educational background, and industry-specific expertise. For instance, an oil painting artist should know drawing principles, regulations, materials, and a variety of painting techniques.
    \item Skill: Skill assessments are designed to measure an individual’s abilities and knowledge in a specific area. Hard skills are specialized, teachable abilities related to a job, such as research or computer. Soft skills include leadership and teamwork, as well as interpersonal skills. Skill is a learned behavior. It is quantifiable and measurable. It can be developed through training and experience.
    \item Ability: Abilities refer to unique characteristics and inherent capabilities that contribute to performing tasks or roles effectively. An ability is innate or natural. It isn't easy to quantify or measure. It is acquired without formal instructions. 
    \item Other: Others were phrases that belonged to neither category above. These were usually rules or requirements described in the job postings. 
\end{itemize}

Once the data was classified into KSAOs, a thorough review was conducted to manually group common terms. This process aimed to identify variations in how similar concepts were expressed across different entries. To standardize the dataset and facilitate easier analysis, a new column was added for each group. This column included a common term representing the variations along with a clear definition of that term. By adopting this approach, we ensured consistency and clarity when managing different representations of the same skills. Table~\ref{tab:skill_variations_examples}, shows examples of the dataset's structure. 

\begin{table*}[!hbtp]
\centering
\caption{Example from the skill variations dataset after grouping of the taxonomy.\label{tab:skill_variations_examples}}
\resizebox{\textwidth}{!}{%
\begin{tabular}{@{}p{3.5cm} p{3.5cm} p{5.5cm} p{2cm}@{}}
\toprule
\textbf{Key} & \textbf{Variation (s)} & \textbf{Definition} & \textbf{Label} \\ \midrule

Actitud positiva (Positive attitude) & Habilidad para mantener una actitud positiva (Ability to maintain a positive attitude) & Intrinsic motivation is the ability to overcome challenges, perform better, and gain satisfaction and enjoyment in the workplace. & Ability \\ 

Excelente nivel de expresion oral y escrita (Excellent level of oral and written expression) & Comunicacion (Communication) & Effectively communicate ideas and thoughts through written or spoken language. & Skill \\ 

Calidad (Quality) & Normas de Calidad (Quality Standards) & Knowledge regarding the quality and testing operations related to manufacturing and different parts of the operations in the automotive industry. & Knowledge \\ 

\bottomrule
\end{tabular}%
}

\end{table*}

\subsection{Inter-annotator agreement}~\label{Section: Agreement}

The inter-annotator agreement for the skill classification task was quantified using Cohen's kappa, resulting in a kappa value of 0.82. This high kappa value reflects a strong agreement beyond chance, underscoring the robustness of the classification process carried out by the annotators.

For the task of skill identification, the annotator reviewed each skill extracted from the job descriptions, validating its relevance and correctness within the context of the dataset. We got a 70\% overlap between the extracted skills and the skills mentioned in the job descriptions that were an exact extraction from the text. However, when focusing on identifying soft skills specifically, the overlap percentage dropped to 52\%. This lower rate can be attributed to the large language models extracting more implicit and nuanced soft skills. We intend to use our gold labels in the future for developing a finetuned version of the skill identification model. 

\subsection{Data Statistics}

Tables~\ref{tab: extracted_jobs_stats} and~\ref{tab:skill_stats} show the detailed statistics of the two datasets. Table~\ref{tab: extracted_jobs_stats} presents statistics on job postings from three major companies: BMW, General Motors, and Tesla. The table includes the number of unique job postings, the total number of jobs, and the counts of unique soft and hard skills identified. BMW has the highest number of unique job postings (69) and total job postings (75), with 463 unique soft skills and 863 unique hard skills identified. General Motors, with 30 unique job postings and the same number of total postings, shows 288 unique soft skills and 415 unique hard skills. Tesla has 41 unique job postings and a total of 45 postings, with 286 unique soft skills and 418 unique hard skills. These statistics highlight the diversity and volume of skills required across different companies and job postings. Table~\ref{tab:skill_stats} provides a distribution of the unique number of skills categorized into knowledge, skills, abilities, and other requirements. The dataset contains a total of 8,484 unique skills, distributed as follows: 2,385 knowledge-related skills, 5,141 general skills, 365 abilities, and 593 other requirements. This distribution illustrates the wide range of competencies and qualifications identified in the job postings, emphasizing the importance of a diverse skill set in the job market.

\begin{table}[!hbtp]
\caption{The table shows the statistics of the subset of the data used for evaluation of skill extraction.\label{tab: extracted_jobs_stats}}
\begin{tabular}{@{}lllll@{}}
\toprule
Companies & \begin{tabular}[c]{@{}l@{}}No. of\\ unique jobs\end{tabular} & \begin{tabular}[c]{@{}l@{}}Total number\\ of jobs\end{tabular} & \begin{tabular}[c]{@{}l@{}}Unique\\soft skills\end{tabular} & \begin{tabular}[c]{@{}l@{}}Unique\\hard skills\end{tabular} \\ \midrule
BMW            &        69            &      75                &  463                                                                          &  863                                                                         \\
General Motors &     30               &        30              & 288                                                                           &        415                                                                   \\
Tesla          &     41               &        45              &  286                                                                          &    418                                                                       \\

\bottomrule
\end{tabular}%

\end{table}

\begin{table}[!hbtp]
\caption{The table shows the unique number of skills distributed by knowledge, skills, abilities and requirements before and after grouping of the taxonomy.\label{tab:skill_stats}}
\begin{tabular}{@{}ccc@{}}
\toprule
             & Distribution & Distribution after grouping \\ \midrule
Knowledge    &     5051    &    134      \\
Skill        &     1227  &    31       \\
Abilities    &     441     &  28          \\
Others &     425       & 27          \\ \hline
Total        &       7144    &     220      \\ \bottomrule
\end{tabular}%

\end{table}

\section{Experimental Design, Materials, and Methods}

\begin{figure*}[ht]
    \centering
    \includegraphics[width=1\linewidth]{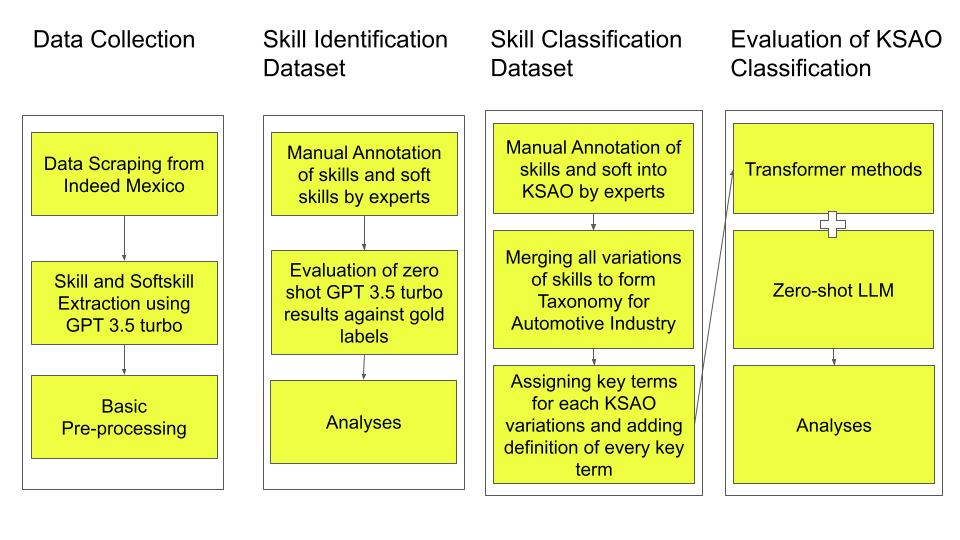}
    \caption{High-level overview of the methodology}
    \label{fig:methedology}
\end{figure*}

The high-level diagram of the methodology can be seen in Figure~\ref{fig:methedology}.

\subsection{Skill Classification}
Following the data acquisition phase, we employed GPT-3.5-turbo in a zero-shot setting to extract both technical skills and soft skills from the job descriptions. GPT-3.5-turbo was leveraged for its ability to comprehend and interpret textual data, thereby facilitating the extraction process seamlessly. Many of the extracted skills are similar to each other but written differently in the job descriptions adding variety to the corpus. We kept them like that so that our model learns to identify them in different ways upon training. Once the relevant skills were identified by the human annotator, we merged all the unique skills to get the final skills. These skills were then annotated into one of the following classes: knowledge, skill, abilities, or others. Following this classification, common terms were grouped together, with each group assigned a key term and a corresponding definition.

\subsubsection{Pre-processing}

Initially, all extracted skills underwent a process of punctuation removal. Then to address duplicate entries effectively, a lemmatization-based approach was adopted. Each skill token was transformed into its corresponding lemma form. This lemma conversion facilitated the identification of exact matches among the extracted skills. Any duplicate entries identified through exact lemma matches were subsequently removed from the dataset.

\subsubsection{Baselines}
We established transformer baseline performance using two state-of-the-art transformer models mBERT~\cite{kenton2019bert} and BETO (BERT-base)~\cite{CaneteCFP2020} on a Spanish language dataset for skill classification. The models were applied to two experimental setups. The first experiment used as input text only, the skill text extracted from job descriptions, while the second incorporated both the skill text and the definition of the common term associated with each skill. 

BETO is a BERT model trained on a big Spanish corpus. BETO is of a size similar to a BERT-base and was trained with the Whole Word Masking technique. The mBERT model on the other hand is a transformer model that has been pretrained on a vast multilingual dataset using a self-supervised learning approach. This means it was trained exclusively on raw text without any human-provided labels. We implemented the base versions of these models and conducted experiments using 5-fold cross-validation. The hyperparameters for both models were set as follows: number of training epochs was 15, learning rate was 3e-5, and the maximum sequence length was set to 100. The performance of each model was assessed based on various metrics, including accuracy, weighted precision, weighted recall, weighted F1 score, macro precision, macro recall, and macro F1 score. 

\begin{table*}[ht]
    \caption{Performance of all models on the Spanish language dataset for skill classification. ACC, Pre, and Rec signify Average Accuracy, Precision, and Recall values, whereas, W and M, refer to Average Weighted and Macro distributions.\label{tab:results}}
\resizebox{\textwidth}{!}{%
    \centering
    \begin{tabular}{llccccccc}
        \hline
   Method  & Model & Acc & Pre\_W & Rec\_W & F1\_W & Pre\_M & Rec\_M & F1\_M \\
        \hline 
     \multirow{4}{*}{Fine tuning}    & BETO & 0.78 & 0.77 & 0.78 & 0.77 & 0.59 & 0.56 & 0.57 \\
       & BETO (with definitions) & 0.85 & 0.87 & 0.85  & 0.84 & 0.83 & 0.78 & 0.77   \\
       & mBERT & 0.75 & 0.74 & 0.75 & 0.75 & 0.56 & 0.53 & 0.54 \\
       & mBERT (with definitions) & 0.86 & 0.91 & 0.86 & 0.87 & 0.82 & 0.80 & 0.79 \\
  \multirow{4}{*}{Zero Shot}    & Mixtral-8x7B-Instruct &  0.48 & 0.66  &  0.48 & 0.52  & 0.36 & 0.39  & 0.34 \\ 
     & Mixtral-8x7B-Instruct
      (with definitions) & 0.38  & 0.66 & 0.38 & 0.43 & 0.34 & 0.37  & 0.29 \\ 
     & GPT-4o mini & 0.40  & 0.70 & 0.40  & 0.44  & 0.37 & 0.39 & 0.31  \\ 
     & GPT-4o mini (with definitions) &  0.38 & 0.70 & 0.38  & 0.41 & 0.36 & 0.39 & 0.30 \\ \hline

    \end{tabular}
    }

\end{table*}

We conducted additional zero shot experiments using two Large Language Models, GPT-4o-mini and Mixtral-8x7B using both experimental setups. GPT-4o-mini serves as a streamlined variant of the advanced ChatGPT-4 model, utilizing a transformer architecture that is optimized for deep conversational understanding. Similar to its predecessor, GPT-4o-mini is designed to facilitate coherent and contextually relevant dialogue during extended interactions. On the other hand, Mistral 8x7B signifies a substantial advancement in the realm of natural language processing, particularly regarding semantic comprehension and generation capabilities. Both models enhance conversational depth, expand semantic understanding, and improve multilingual capabilities~\cite{ono_2024evaluating}.

To ensure deterministic responses for both models, the temperature was set to 0, while the response format followed the structure of the KSAO. The system message instructed the models to classify the input text into one of the four categories: knowledge, skill, ability, or others, based on the definitions provided in Section 3.3.2 for each class.

\section{Results}\label{Section: Results}

The results of our KSAO classification experiments, summarized in Table~\ref{tab:results}, reveal that both BETO and mBERT models performed well in the task of skill classification on the Spanish language dataset, with some notable differences. The BETO model, when fine-tuned, achieved an accuracy of 0.78, with weighted precision, recall, and F1 scores of 0.77, 0.78, and 0.77, respectively. At the macro level, BETO’s precision, recall, and F1 scores were 0.59, 0.56, and 0.57.

The inclusion of definitions in the BETO model significantly boosted its performance. With definitions, BETO achieved an accuracy of 0.85, with weighted precision, recall, and F1 scores of 0.87, 0.85, and 0.84, respectively. Its macro precision, recall, and F1 scores improved to 0.83, 0.78, and 0.77, indicating better overall performance when definitions were utilized during fine-tuning.

In comparison, the mBERT model achieved slightly lower results. Fine-tuning mBERT without definitions resulted in an accuracy of 0.75, with weighted precision, recall, and F1 scores of 0.74, 0.75, and 0.75, respectively. The macro-level scores for mBERT were also slightly lower than BETO, with precision, recall, and F1 scores of 0.56, 0.53, and 0.54. However, with definitions, mBERT’s performance surpassed BETO, achieving an accuracy of 0.86, weighted precision of 0.91, weighted recall of 0.86, and a weighted F1 score of 0.87. The macro-level performance also showed improvement with definitions, reaching precision, recall, and F1 scores of 0.82, 0.80, and 0.79, respectively.

In the zero-shot setting, all models performed considerably worse than in the fine-tuning setting. The Mixtral-8x7B-Instruct model achieved an accuracy of 0.48 and a weighted F1 score of 0.52. When definitions were included, Mixtral’s performance dropped, with accuracy and weighted F1 scores of 0.38 and 0.43, respectively. At the macro level, the inclusion of definitions also resulted in lower precision, recall, and F1 scores. Similarly, the GPT-4o mini model in the zero-shot setting achieved an accuracy of 0.40 and a weighted F1 score of 0.44. Including definitions slightly lowered its accuracy to 0.38 and its weighted F1 score to 0.41, with a similar decline in macro-level metrics.

The models' performance compared to inter-annotator agreement is noteworthy. Both BETO and mBERT, particularly when fine-tuned with definitions, achieved higher accuracy than the typical agreement between human annotators. This suggests that these models were able to generalize well and capture patterns that may be inconsistently recognized by human annotators. This discrepancy highlights the advantage of models that consistently apply learned rules and are less prone to the subjectivity and errors associated with human annotators, such as fatigue or variation in interpretation.

In the skill identification task, the model’s performance was evaluated using a human evaluation process, as explained in Section 3.4. The results of this evaluation showed a 70\% overlap between the skills extracted by the model and those explicitly mentioned in the job descriptions, indicating that the model was generally effective at identifying skills directly stated in the text. However, when the focus shifted to identifying soft skills specifically, the overlap percentage dropped to 52\%. This lower rate can be attributed to the large language models' ability to extract more implicit and nuanced soft skills, which are often not explicitly mentioned but inferred from the job descriptions.

While the model shows promise, there is room for improvement, particularly in the accurate identification of soft skills. This observation highlights the challenge of soft skill extraction, as these skills tend to be less concrete and harder to detect in a straightforward manner. Despite the lower overlap, the model’s ability to infer such implicit skills is valuable for enhancing job description analysis. To further improve the accuracy of soft skill identification, we plan to use our human-annotated gold labels for developing a fine-tuned version of the skill identification model. This future work aims to refine the model’s ability to capture both explicit and implicit skills, particularly focusing on the nuanced detection of soft skills in various job descriptions.

\section{Error Analyses}~\label{Section: Error Analyses}

One common error is the misclassification of an example as knowledge when it should be categorized under rules and requirements, and vice versa. For instance \textit{"Proficiency in Python programming"} can be correctly identified as knowledge. However, \textit{"Adherence to Python coding standards"} is sometimes mistakenly classified as a knowledge instead of a rule or requirement. The former indicates information and experience, while the latter specifies a guideline that must be followed. Similarly, distinguishing between skills and knowledge presents challenges. The subtle differences in context can lead to misclassifications i.e. \textit{"Understanding of financial principles"} is a knowledge item, as it reflects theoretical understanding. However, it can sometimes be misclassified as a skill. In contrast, \textit{"hability to learn and acquire new knowledge"} is a skill, as it demonstrates the capacity to gain new information and skills. These examples highlight the nuanced distinctions between skills, abilities, rules and requirements, and knowledge. The context in which terms are used significantly impacts their classification. Improvements in model training, including more context-aware processing and finer granularity in annotation guidelines, could help reduce these types of errors. 

The confusion matrix shown in Table ~\ref{tab:confusion_martix_mbert} illustrates the average performance of the mBERT classification model with definitions. Among all the models tested, this model demonstrated the strongest overall performance. The Knowledge category, in particular, achieved the highest number of correct classifications, highlighting the model’s effectiveness in this area. However, this category also had the highest number of instances, suggesting a potential class imbalance that may need to be addressed.

\begin{table}[]
\caption{Average Confusion Matrix Results for mBERT Model with Definitions. X-axis represents the predicted values, while the y-axis represents the real values.\label{tab:confusion_martix_mbert}}
\begin{tabular}{@{}lllll@{}}
\hline
 & Skill & Ability & Knowledge & Others \\ \hline
Skill & 188.8 & 29.4 & 27.4 & 0.0 \\
Ability & 19.0 & 61.8 & 3.2 & 3.8 \\
Knowledge & 84.6 & 3.0 & 919.0 & 17.4 \\
Others & 1.8 & 2.2 & 8.6 & 72.6 \\ \bottomrule
\end{tabular}%

\end{table}

Despite the strong results in Knowledge, there is considerable confusion between the Skill and Knowledge categories, with misclassifications occurring in both directions. This overlap suggests that these categories may share some similarities or features that the model finds difficult to distinguish. On the other hand, the Others category is relatively well-distinguished, with only minor confusion, mostly with Knowledge.

A key recommendation for future work is to address the class imbalance present in the dataset. The disproportionate number of instances in the Knowledge category highlights the need for techniques such as oversampling, undersampling, or adjusting class weights to ensure a more balanced dataset. Prioritizing the improvement of the classification of Ability, which currently has the lowest correct classification rate, is also crucial. Additionally, enhancing the model’s ability to distinguish between similar categories, such as Skill and Knowledge, and refining its accuracy in classifying Ability, would lead to significant improvements in overall performance.

\section{Ethical Considerations}
The deployment of AI models, particularly in sensitive areas such as skill identification and classification, must be done responsibly. We emphasize that the models developed in this study are intended to support and enhance human decision-making, not replace it. Users should be aware of the models' limitations and use the results as part of a broader, human-in-the-loop decision-making process. We advocate for continuous monitoring and evaluation of the models in real-world applications to ensure they perform as intended and do not produce unintended harmful effects.

\section{Conclusion}
In conclusion, this paper addressed the lack of comprehensive Spanish language datasets for skill identification and classification tasks. We developed a novel dataset containing job descriptions from major automotive companies in Mexico, annotated for skills as well as their classification into knowledge, skills, abilities, and other requirements based on the KSA taxonomy. Our detailed annotation methodology and guidelines contribute to a standardized approach for distinguishing between these different competency categories. Using deep learning baselines of mBERT and BETO models, we demonstrated that transformer architectures can achieve high accuracy in automating the skill classification task on this Spanish dataset.  However, error analysis revealed some nuanced challenges in differentiating skills from knowledge items or distinguishing \say{other} class (rules/requirements), often dependent on linguistic context. Moving forward, the created dataset can serve as a valuable resource for further research and deployment of skill extraction and categorization systems for the Spanish-speaking job market.

\section{Acknowledgments}
I would like to express my profound gratitude to Santander Group (Banco Santander S.A.) for their generous sponsorship, which has been instrumental in the successful completion of this research.


\bibliography{main.bib}

\end{document}